\documentclass{article}

\usepackage{arxiv}

\usepackage[utf8]{inputenc} 
\usepackage[T1]{fontenc}    
\usepackage{hyperref}       
\usepackage{url}            
\usepackage{booktabs}       
\usepackage{amsfonts}       
\usepackage{nicefrac}       
\usepackage{microtype}      

\usepackage{url,hyperref,lineno,microtype,subcaption}
\usepackage[onehalfspacing]{setspace}
\usepackage{booktabs}
\usepackage{caption}
\usepackage{tabularx}
\usepackage{rotating}
\usepackage{caption}
\usepackage{array}
\usepackage{amsmath}
\usepackage{graphicx,verbatim}
\usepackage[ruled,vlined]{algorithm2e}

\title{SlideMamba: Entropy-Based Adaptive Fusion of GNN and Mamba for Enhanced Representation Learning in Digital Pathology}

\author{
 Shakib Khan \\
  Computational Science and Informatics\\
  Roche Diagnostic Solutions\\
  Pathology Lab, Canada \\
   \And
 Fariba Dambandkhameneh \\
  Computational Science and Informatics\\
  Roche Diagnostic Solutions\\
  Pathology Lab, Canada \\
  \texttt{fariba.dambandkhameneh@roche.com} \\
  \And
 Nazim Shaikh \\
  Computational Science and Informatics\\
  Roche Diagnostic Solutions\\
  Pathology Lab, USA \\
  \And
 Yao Nie \\
  Computational Science and Informatics\\
  Roche Diagnostic Solutions\\
  Pathology Lab, USA \\
   \And
 Raghavan Venugopal \\
  Computational Science and Informatics\\
  Roche Diagnostic Solutions\\
  Pathology Lab, USA \\
   \And
 Xiao Li \\
  Computational Science and Informatics\\
  Roche Diagnostic Solutions\\
  Pathology Lab, USA \\
  \texttt{xiao.li.xl2@roche.com} \\ 
}

\begin{document}
\maketitle
\begin{abstract}
Advances in computational pathology increasingly rely on extracting meaningful representations from Whole Slide Images (WSIs) to support various clinical and biological tasks. In this study, we propose a generalizable deep learning framework that integrates the Mamba architecture with Graph Neural Networks (GNNs) for enhanced WSI analysis. Our method is designed to capture both local spatial relationships and long-range contextual dependencies, offering a flexible architecture for digital pathology analysis. Mamba modules excels in capturing long-range global dependencies, while GNNs emphasize fine-grained short-range spatial interactions. 
To effectively combine these complementary signals, we introduce an adaptive fusion strategy that uses an entropy-based confidence weighting mechanism. This approach dynamically balances contributions from both branches by assigning higher weight to the branch with more confident (lower-entropy) predictions, depending on the contextual importance of local versus global information for different downstream tasks.
We demonstrate the utility of our approach on a representative task: predicting gene fusion and mutation status from WSIs. Our framework, SlideMamba, achieves an area under the precision recall curve (PRAUC) of 0.751 ± 0.05, outperforming MIL (0.491 ± 0.042), Trans-MIL (0.39 ± 0.017), Mamba-only (0.664 ± 0.063), GNN-only (0.748 ± 0.091), and a prior similar work GAT-Mamba (0.703 ± 0.075). SlideMamba also achieves competitive results across ROC AUC (0.738 ± 0.055), sensitivity (0.662 ± 0.083), and specificity (0.725 ± 0.094). These results highlight the strength of the integrated architecture, enhanced by the proposed entropy-based adaptive fusion strategy, and suggest promising potential for application of spatially-resolved predictive modeling tasks in computational pathology.
\end{abstract}


\section*{Introduction}

The analysis of Whole Slide Images (WSIs) plays an increasingly vital role in computational pathology, enabling scalable, image-based alternatives to molecular assays for a wide range of clinical and biological tasks. From cancer subtyping to biomarker discovery and genetic variant prediction, deep learning approaches have demonstrated the potential to uncover morphological patterns indicative of underlying molecular alterations. However, although prior work has highlighted the importance of quantifying such spatial relationships across scales \cite{li2024deciphering},
effective modeling of WSIs remains challenging due to the need to capture both fine-grained local interactions, such as tumor-stroma boundaries or immune cell infiltration, and broad tissue-level context, such as global architectural organization across centimeter-scale slides.

Existing approaches often prioritize one end of this spectrum at the expense of the other. For example, Graph Neural Networks (GNNs) have shown promise in addressing short-range dependencies by modeling WSIs as graphs, where nodes represent regions of interest (ROIs) and edges encode spatial or morphological relationships between adjacent tissue regions \cite{lu2022slidegraph+}. This approach excels at capturing localized patterns, such as tumor-stroma interactions, but often overlooks global tissue context. Conversely, state space models (SSMs), exemplified by the recently proposed Mamba architecture, offer linear computational complexity and dynamic feature selection, enabling efficient learning of long-range dependencies in high-resolution images \cite{gu2023mamba}. Vision Mamba, adapted for imaging tasks, has demonstrated superior performance in capturing global context compared to ViTs, while maintaining computational efficiency \cite{liu2024vision}. 
While recent work by Ding et al. \cite{ding2024combining} attempted to combine GNN and Mamba through element-wise summation of short- and long-range embeddings, this simple additive approach implicitly assumes that local and global representations contribute equally and consistently across all samples. In practice, however, the relative importance of short- and long-range dependencies can vary greatly between tissue types, histological patterns, and clinical tasks. Rigid summation fails to account for this dynamic variation, potentially diluting critical information when one source is more informative than the other. Moreover, naive summation may introduce conflicting signals if the two branches generate embeddings with different statistical properties or scales, making the fused representation suboptimal. These limitations underscore the need for a more flexible and data-driven fusion mechanism that can adaptively weight each branch according to its confidence and contextual relevance.

To address this gap, we present \textbf{SlideMamba}, a general-purpose framework for WSI analysis that adaptively fuses GNN and Mamba representations to capture both short- and long-range dependencies. Unlike prior fixed fusion strategies, SlideMamba introduces an entropy-based confidence weighting mechanism that dynamically adjusts the contributions from the local (GNN) and global (Mamba) branches. By assigning higher weights to the branch with more confident (lower-entropy) predictions, SlideMamba learns to emphasize the most reliable source of information for each input, enabling robust multi-scale integration tailored to diverse downstream tasks.

We demonstrate the utility of SlideMamba on the representative task of predicting gene fusions and mutations directly from WSIs. While this application serves as a clinically relevant test case, our proposed framework is broadly applicable to other downstream tasks in computational pathology. Across multiple evaluation metrics, SlideMamba outperforms both GNN-only baseline (SlideGraph+\cite{lu2022slidegraph+}) and the Mamba-only baseline, as well as other competitive approaches including MIL\cite{ilse2020deep}, TransMIL\cite{shao2021transmil}, and GAT-Mamba\cite{ding2024combining}, highlighting the advantages of adaptive multi-scale integration.

The remainder of this manuscript is organized as follows: Section 2 reviews Related Work, critically analyzing existing methods for modeling short-range (GNNs) and long-range (Transformers, Mamba) dependencies in computational pathology, with emphasis on their limitations in multi-scale integration. Section 3 details our Proposed Methodology, introducing the SlideMamba framework and its entropy-based adaptive fusion. Section 4 presents experimental validation through a representative case study predicting gene fusions and mutations from WSIs using an in-house clinical trial dataset, illustrating the practical application of SlideMamba. Finally, Section 5 discusses clinical implications, limitations, and future directions for adaptive multi-scale modeling in digital pathology.

\section*{Related Work}
\subsection*{Graph-based approaches in computational pathology}
In recent years, GNNs have achieved remarkable success across various domains, including biology, drug discovery~\cite{gaudelet2021utilizing}, disease prediction, and biomedical imaging~\cite{wu2020comprehensive}. One promising application of GNNs is in computational pathology, which requires strong spatial and relational reasoning. Whole Slide Images (WSIs) are increasingly represented as graphs~\cite{konda2020graph}, where tissue regions serve as nodes and their spatial relationships form the edges. By employing message-passing GNNs, it becomes possible to effectively capture spatial connectivity and interactions between tissue regions, enabling more accurate and interpretable analyses of WSIs.

To enhance the modeling capabilities of GNNs, several advanced architectures have been introduced, including Graph Convolutional Networks (GCN)~\cite{kipf2016semi}, Graph Attention Networks (GAT)~\cite{velivckovic2017graph}, Graph Isomorphism Networks (GINConv)~\cite{xu2018how}, EdgeConv~\cite{wang2019dynamic}, and GraphSAGE~\cite{hamilton2017inductive}. These models leverage message-passing mechanisms to iteratively aggregate information from neighboring nodes, refining node representations over multiple layers. While each type of GNN employs distinct aggregation and update functions, they all contribute to capturing different structural characteristics within the graph, improving performance across various tasks. Building on these advancements, SlideGraph+~\cite{lu2022slidegraph+} employs a graph convolutional architecture, primarily using EdgeConv layers, to model spatial relationships between tissue regions at WSI scale.
This approach proved effective in predicting HER2 status directly from H\&E-stained breast cancer (BCa) tissue slides, demonstrating the potential of GNN-based methods in computational pathology. 
Similarly, NAGCN~\cite{guan2022node} addresses the challenge of gigapixel WSIs by constructing a node-aligned, hierarchical graph representation. The method uses a global clustering operation to establish consistent patch-level correspondence across slides, followed by local cluster–based sampling within each WSI, enabling the model to capture both global tissue distribution and fine-grained local structure. Applied to cancer subtype classification, NAGCN outperforms standard WSI graph methods, demonstrating improved alignment and representational fidelity.

Despite these advances, GNNs are not without limitations. One significant drawback is their susceptibility to the over-smoothing issue~\cite{topping2022understanding}, where node representations tend to become indistinguishable after multiple iterations of neighborhood aggregation, ultimately degrading the model’s capacity to capture meaningful local variation. Additionally, traditional message-passing mechanisms inherently restrict information propagation to immediate or near-local neighborhoods, making it challenging for GNNs to effectively capture long-range dependencies between spatially distant tissue regions within gigapixel WSIs~\cite{alon2021on}. This locality bias can lead to incomplete modeling of global tissue architecture and cross-region interactions that are critical in many pathology tasks. Furthermore, as the size of the graph grows, which is typical for WSIs with thousands of patches, the computational and memory demands of message passing scale poorly, posing practical challenges for training and inference. These limitations highlight the need for novel architectural modifications or hybrid designs that can preserve the strengths of GNNs in local context modeling while extending their reach to global structural patterns in a computationally efficient manner.

\subsection*{Sequence-based approaches in computational pathology}
While graph-based models excel at capturing local spatial interactions within Whole Slide Images (WSIs), effectively modeling long-range dependencies remains a challenge due to the locality constraints of message passing. To address this gap, sequence-based models, particularly those leveraging self-attention mechanisms, have gained prominence in computational pathology.

Transformers~\cite{vaswani2017attention}, initially developed for natural language processing, have been successfully adapted to computer vision tasks through the Vision Transformer (ViT)~\cite{dosovitskiy2021an}. ViT and its derivatives excel at capturing long-range dependencies by leveraging self-attention to learn pairwise relationships across all image patches, making them highly effective for modeling global context. In WSI analysis, this ability to capture long-range interactions is especially valuable for tasks such as cancer subtyping and molecular status prediction, where spatially distant regions may exhibit correlated morphological patterns. Building on this foundation, TransMIL~\cite{shao2021transmil} extends the Transformer architecture to the multiple instance learning (MIL)~\cite{ilse2020deep} setting, which is well-suited for gigapixel WSIs with weak slide-level labels. By applying self-attention to sets of image patches (instances), TransMIL effectively models inter-patch relationships and aggregates relevant contextual information for robust slide-level inference. Further refinements, such as ACMIL~\cite{zhang2024attention}, introduce multiple attention branches to better identify and combine diverse discriminative regions, enhancing the model’s capacity to capture the heterogeneous structures typical of histopathological slides.

Despite their effectiveness, Transformer-based approaches face practical constraints when applied to large WSIs. The quadratic computational and memory complexity of self-attention can become prohibitive as the number of patches increases, limiting scalability for high-resolution whole-slide analysis. To address these challenges, recent developments have explored State Space Models (SSMs) as efficient alternatives for long-range sequence modeling. In this regard, Mamba~\cite{gu2023mamba} represents a notable advance: it replaces the costly pairwise self-attention mechanism with a continuous-time state space formulation that operates in linear time with respect to sequence length. This design enables Mamba to capture long-range dependencies and dynamically select relevant features without incurring the memory overhead typical of Transformers. By decoupling sequence length from computational cost, Mamba achieves superior scalability, making it particularly well suited for gigapixel WSIs composed of tens of thousands of patches. Additionally, Mamba’s flexible state space architecture allows it to adaptively model temporal or spatial correlations, demonstrating strong performance across large-scale vision tasks and indicating significant promise for computational pathology applications where both efficiency and global context modeling are critical.

\subsection*{Hybrid approaches in computational pathology}

To address the inherent limitations of purely graph-based or sequence-based methods, recent studies have explored hybrid architectures that combine Graph Neural Networks (GNNs) with Transformers or State Space Models (SSMs). For example, GAT-Mamba~\cite{ding2024combining} integrates Graph Attention Networks to capture local relational reasoning with the Mamba architecture for efficient modeling of long-range dependencies. While this type of hybrid design represents an important step toward multi-scale representation learning for WSIs, it still suffers from several notable limitations.

First, many existing hybrid models adopt simple element-wise operations (such as summation) to merge local and global representations, implicitly assuming that both branches contribute equally to the final prediction. This rigid fusion strategy fails to adapt to the dynamic variation in contextual relevance of local versus global signals across different tissue regions and cases, potentially diluting critical information when one source is more informative than the other. Moreover, naive summation can introduce conflicting signals if the two branches generate embeddings with different statistical properties, magnitudes, or scales, making the fused representation suboptimal and harder to interpret. In practice, the embeddings produced by the local (GNN) and global (Mamba) branches may not be aligned dimensionally or semantically. For example, channel 1 in the GNN output may encode a spatial boundary feature, while channel 1 in the Mamba output may represent a global architectural pattern. Summing such semantically misaligned channels can lead to incoherent or meaningless feature combinations, ultimately undermining the discriminative power of the fused representation.

Furthermore, most early hybrid models do not incorporate explicit mechanisms for evaluating the relative confidence or uncertainty associated with each branch’s output. In the absence of such mechanisms, noisy or unreliable features from one branch may dominate more informative signals from the other, thereby compromising predictive performance. Moreover, without the capacity to adjust fusion weights dynamically during inference, these models are ill-equipped to accommodate the heterogeneous and intricate tissue structures characteristic of gigapixel whole-slide images

These shortcomings motivate our proposed framework, SlideMamba, which extends the strengths of GNNs and Mamba through an adaptive, entropy-based fusion strategy. By dynamically weighting local and global signals based on prediction confidence, SlideMamba flexibly integrates short- and long-range dependencies while mitigating the risk of noisy, misaligned, or conflicting signals dominating the final representation. This context-aware fusion mechanism provides a robust and scalable solution for large-scale WSI analysis, delivering more reliable multi-scale integration than previous hybrid approaches.

\section*{Methodology}
\subsection*{Graph Construction}
For each WSI, the graph $G$ is constructed by defining nodes and edges through feature extraction from the tiles. Tiles are sampled at 40x magnification to preserve high-resolution morphological details. Feature extraction is performed using the UNI model~\cite{chen2024towards}, a self-supervised learning framework specifically designed for pathology image analysis, which generates 1024-dimensional deep feature representations for each tile. These features encode the semantic content of the tissue regions. In addition, N-dimensional sinusoidal positional encodings, with $N=16$, are computed from the relative spatial coordinates of each tile within the WSI. These encodings provide spatial context, enabling the model to capture the positional relationships among tiles.

For edge features, two key components are considered: (1) the cosine similarity between the UNI-extracted deep features of connected tiles, and (2) the Euclidean distance between their spatial coordinates. These features help encode both semantic and spatial relationships between nodes. To define the edge connectivity in the graph, the k-nearest neighbors (k-NN) approach is utilized, with $k = 8$. The Euclidean distance between tiles serves as the distance metric for determining connectivity. This design is based on the assumption that neighboring tiles share similar contextual information, thereby enabling the model to capture local interactions and dependencies effectively~\cite{chen2021whole}.

\subsection*{Model Architecture}
\begin{sidewaysfigure}
    \centering
    \includegraphics[width=1\textwidth]{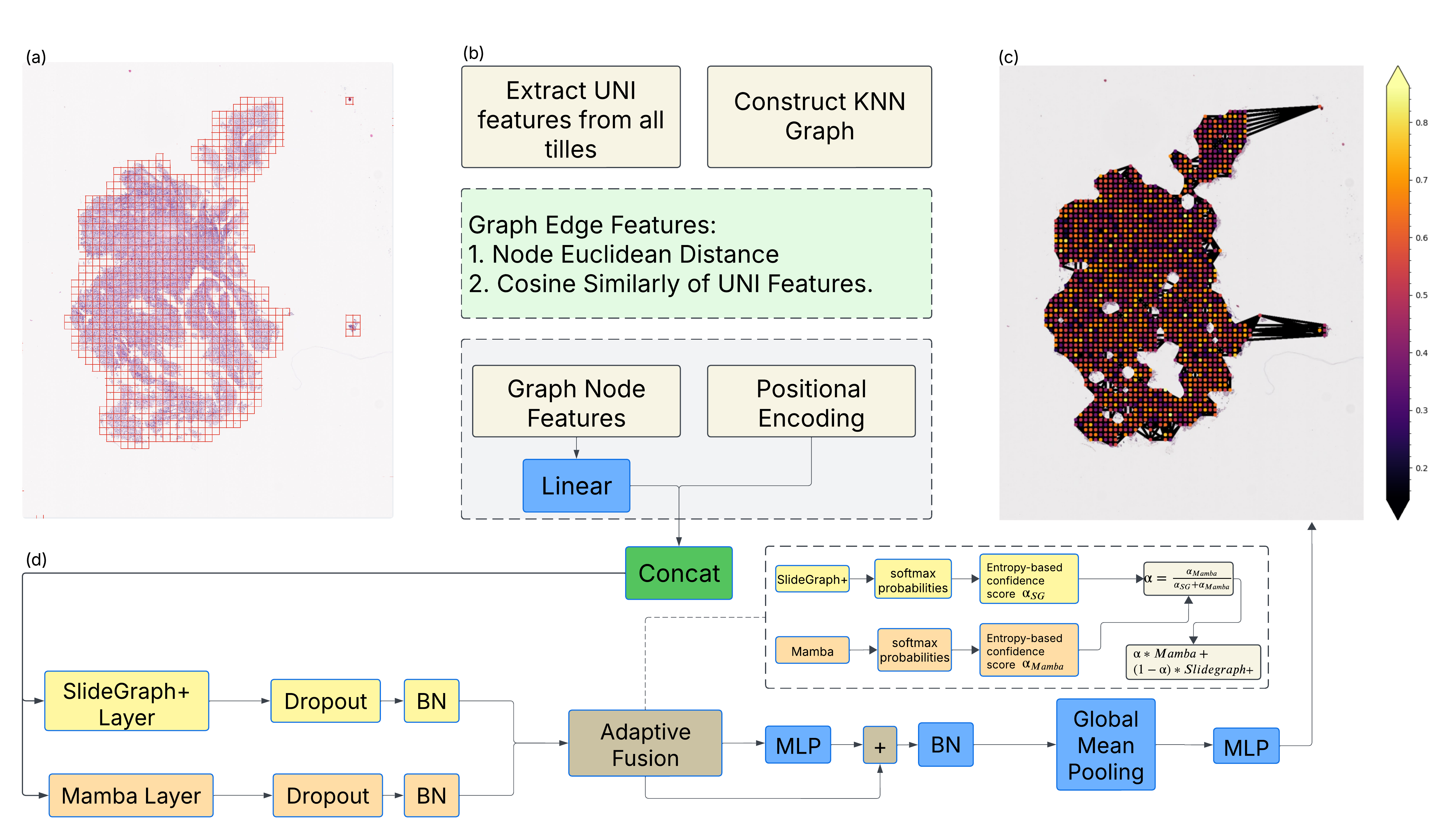}
    \caption{The proposed SlideMamba pipeline comprises several key steps: (a) tiling the whole-slide image (WSI) into patches; (b) extracting node and edge features for each patch, with positional encodings incorporated to enrich spatial context; (c) constructing a graph representation of the WSI; and (d) performing modeling with two complementary branches — one capturing local dependencies and the other modeling global dependencies. The outputs from these two branches are fused adaptively through an entropy-based confidence weighting mechanism, which assigns higher weights to the branch with more confident (low-entropy) predictions. The final fused representation is processed by batch normalization (BN), a multi-layer perceptron (MLP), and global mean pooling to enable robust slide-level inference.}
    \label{SlideMamba}
\end{sidewaysfigure}

Figure~\ref{SlideMamba} illustrates the workflow of the SlideMamba pipeline. In this approach, each WSI is first divided into non-overlapping tiles of uniform size, standardized to $224\times224$ pixels at 40× magnification to ensure consistent resolution and area coverage across all slides, regardless of variations in slide preparation or scanning parameters. This standardization is critical for maintaining uniformity in feature extraction for both the graph-based and sequence-based branches. In the graph-based branch, the WSI is modeled as a graph in which nodes represent individual tissue regions and edges capture their spatial and semantic relationships. In parallel, the sequence-based branch models the tile embeddings, which are augmented with positional encodings, as an ordered sequence to capture long-range contextual dependencies. Together, these complementary representations provide a comprehensive characterization of local and global tissue structures.

\subsubsection*{SlideGraph+:} 
\noindent GNNs have demonstrated exceptional performance in capturing local dependencies and spatial relationships within structured data, such as tissue regions in WSIs. GNNs achieve this through the message-passing mechanism, which iteratively aggregates information from neighboring nodes and updates each node’s features. This process enables GNNs to effectively model the spatial connectivity and interactions between distinct tissue regions. Different types of GNNs employ unique aggregation and update functions, allowing them to capture diverse aspects of graph structure.
In this work, we build on the SlideGraph+ (SG) framework~\cite{lu2022slidegraph+} to extract short-range information from WSIs. While the original SlideGraph+ primarily uses EdgeConv layers to model spatial relationships, we adapt its backbone to instead employ GINConv (Graph Isomorphism Network)~\cite{xu2018powerful} due to its stronger capacity to capture fine-grained local structural patterns. GINConv is particularly well-suited for this purpose because it is theoretically as powerful as the Weisfeiler-Lehman graph isomorphism test, allowing it to distinguish between subtle differences in graph structures more effectively than many other GNN variants. 
This makes GINConv well-suited for modeling the complex and heterogeneous relationships between tissue regions in WSIs. The following describes how GINConv updates its node features:

\begin{equation}
h_{v}^{(k)} = \text{MLP}^{(k)} \left( \left( 1 + \epsilon^{(k)} \right) \cdot h_{v}^{(k-1)} + \sum_{u \in N(v)} h_{u}^{(k-1)} \right).
\label{GINConv}
\end{equation}
where $h_{v}^{(k)}$ is the feature representation of node $v$ at the $k$-th layer, and $h_{v}^{(k-1)}$ is the feature from the previous layer. The term $\epsilon^{(k)}$ is a learnable parameter that allows the model to adjust the weighting of the node’s own features relative to its neighbors. The summation $\sum_{u \in N(v)} h_{u}^{(k-1)}$ aggregates the feature representations of all neighboring nodes $u$ in the neighborhood $N(v)$. The combined result is then passed through a multi-layer perceptron (MLP) to produce the updated node representation.

\subsubsection*{Mamba:} Similar to Recurrent Neural Networks (RNNs), State Space Models (SSMs) map an input sequence \( x(t) \in \mathbb{R}^N \) to an output sequence \( y(t) \in \mathbb{R}^N \) through a hidden state \( h(t) \in \mathbb{R}^N \), governed by a linear ordinary differential equation for continuous input:
\begin{equation}
\begin{split}
h'(t) &= Ah(t) + Bx(t) \\
y(t) &= Ch(t).
\end{split}
\label{Mamba}
\end{equation}

In this formulation, \(\mathbf{A}\) is the state transition matrix that compresses all historical information, \(\mathbf{B}\) projects the input, and \(\mathbf{C}\) maps the hidden state to the output. The term \( Ah(t) \) describes the evolution of the internal state, \( Bx(t) \) governs the influence of the input, and \( Ch(t) \) translates the latent state into output signals. To operate in discrete time, \(\mathbf{A}\) and \(\mathbf{B}\) are discretized using a step size \(\Delta\):
\[
\overline{A} = \exp(\Delta A), \quad
\overline{B} = (\Delta A)^{-1} (\exp(\Delta A) - I) \Delta B.
\]

Traditional SSMs and their structured variant, the Structured State Space Sequence model (S4)~\cite{gu2021efficiently}, can capture long-range dependencies by encoding past information in a compact latent state. However, their time-invariant nature limits context-awareness: the parameters \(\mathbf{A}\), \(\mathbf{B}\), and \(\mathbf{C}\) do not change with the input sequence, which constrains adaptability for complex or highly variable signals. Mamba~\cite{gu2023mamba} addresses this limitation by introducing an input-dependent selection mechanism. In Mamba, the projection parameters \(\mathbf{B}\) and \(\mathbf{C}\), as well as the discretization step size \(\Delta\), become functions of the input, enabling the model to adaptively propagate or filter information at each step based on local context. This allows Mamba to dynamically focus on salient parts of the sequence and suppress irrelevant signals, which is crucial when modeling long sequences such as patch-level features in whole-slide pathology images. Unlike conventional SSMs and S4, whose time-invariant structure enables full parallelization through convolution, Mamba’s time-varying parameters prevent simple convolutional computation. To overcome this, Mamba employs a hardware-efficient parallel scan algorithm, which reparameterizes the recurrent computation into an operation that can be parallelized across time steps. Combined with kernel fusion and recomputation techniques, this design delivers high computational efficiency for large-scale sequence modeling.

In this work, each patch or node-level feature extracted from the WSI graph is treated as a token in the input sequence. Mamba’s selection mechanism dynamically downweights less relevant patches while preserving long-range spatial context. However, once graph-structured patches are linearized for sequential modeling, explicit graph connectivity is lost. To compensate, we incorporate sinusoidal positional encodings, similar to those used in vision Transformers, to ensure that the relative relationships between tissue regions are preserved during sequence-based processing with Mamba.

\subsubsection*{SlideMamba:}

\noindent Building on the strengths of both GNN-based and sequence-based paradigms, we introduce SlideMamba, a unified framework designed to effectively capture both short-range and long-range dependencies within Whole Slide Images (WSIs). The overall workflow is detailed in Algorithm~\ref{alg:SlideMambablock} and conceptually illustrated in Figure~\ref{SlideMamba}.

\noindent The SlideMamba pipeline begins by constructing complementary representations for each WSI to support both branches. For the graph-based branch, each WSI is modeled as a graph where nodes correspond to individual tissue tiles and edges encode their spatial and semantic relationships. Node features are extracted using embeddings from a pretrained UNI~\cite{chen2024towards} model and are further enriched with $N$-dimensional sinusoidal positional encodings to preserve spatial context. Edge features capture pairwise cosine similarity of deep features and Euclidean distances between tile coordinates, with connectivity established via a $k$-nearest neighbors (k=8) approach; In parallel, the same UNI-extracted tile embeddings and positional encodings are arranged as an ordered sequence to serve as input to the Mamba branch. This sequential representation enables the model to capture long-range dependencies across the entire slide by treating each tile as a token in a sequence, leveraging Mamba's efficient state space mechanism for scalable global context modeling.

This dual-branch architecture combines the strengths of SlideGraph+ (SG) for fine-grained local structural modeling and Mamba for efficient long-range sequence modeling. Crucially, SlideMamba employs an entropy-based adaptive fusion mechanism that dynamically weights the contributions of the local and global branches according to their prediction confidence. This ensures that the final slide-level inference remains robust and context-aware, effectively adapting to the heterogeneous tissue patterns often encountered in WSIs.

The core of SlideMamba lies in the \textit{SlideMambaBlock} (Algorithm~\ref{alg:SlideMambablock}), which processes the graph information through two parallel branches:

\begin{itemize}
    \item \textbf{SlideGraph+ Branch}: This branch processes the input node features~$\mathbf{X}$ and edge features~$\mathbf{E}$ together with the adjacency matrix~$\mathbf{A}$ using a SlideGraph+ layer (specifically, GINConv). As formulated in Equation~\ref{GINConv}, GINConv iteratively updates node embeddings by aggregating information from neighboring nodes, effectively modeling local structural patterns and short-range dependencies within the WSI. The resulting representations are regularized using batch normalization and dropout.
    
    \item \textbf{Mamba Branch}: In parallel, the Mamba branch treats the same node features~$\mathbf{X}$ as an ordered sequence. Following the state-space model described in Equation~\ref{Mamba}, Mamba dynamically selects which information to retain or discard, enabling efficient modeling of long-range dependencies across the entire slide. This branch captures global tissue context that is often inaccessible to message-passing GNNs alone. Its output is normalized and regularized, similar to the SlideGraph+ branch.
\end{itemize}

\noindent
\textbf{Entropy-Based Adaptive Fusion.}  

Entropy has been increasingly recognized as a powerful tool for quantifying uncertainty and complexity in digital pathology \cite{li2025entropy}. Inspired by this, to flexibly integrate the complementary strengths of both branches, we introduce a novel entropy-based adaptive fusion mechanism, \textit{EntropyConfidence} (Algorithm~\ref{alg:EntropyConfidence}). Instead of relying on fixed or static weights, this approach adaptively balances local and global representations based on prediction confidence. Specifically, each branch’s softmax output is used to estimate the normalized predictive entropy, where lower entropy indicates higher confidence.

For each branch output, an entropy-based confidence weight is computed as:
$$
w_{\text{SG}} = 1 - H(\hat{\mathbf{y}}_{\text{SG}}), 
\quad 
w_{\text{Mamba}} = 1 - H(\hat{\mathbf{y}}_{\text{Mamba}})
$$
with 
$$H(\hat{\mathbf{y}}) = -\frac{1}{\log(C)} \sum_{c=1}^{C} \hat{y}_c \log(\hat{y}_c).$$
where $C$ is the number of classes and $\hat{\mathbf{y}}$ denotes the softmax probability vector.

The fused representation is then formed by:
\[
\mathbf{X}^{l+1}_{\text{SlideMamba}} 
= (1 - \alpha)\,\mathbf{X}^{l+1}_{\text{SG}} + \alpha\,\mathbf{X}^{l+1}_{\text{Mamba}},
\quad
\text{with} \quad
\alpha = \frac{w_{\text{Mamba}}}{w_{\text{SG}} + w_{\text{Mamba}}}.
\]

This entropy-guided fusion ensures that the branch with more confident predictions (lower entropy) contributes more strongly to the final representation, allowing the model to adaptively emphasize short-range (SG) or long-range (Mamba) context depending on the tissue pattern.

\noindent
The fused features are further refined with an MLP layer and a residual connection, followed by batch normalization, ensuring stable training and effective information flow. After stacking multiple SlideMambaBlocks, a global mean pooling aggregates the node-level representations into a whole-slide representation, which is passed to a final MLP for predicting the slide-level outcome.

\begin{algorithm}
\caption{SlideMambaBlock}
\label{alg:SlideMambablock}
\KwIn{Node features $X_{\text{UNI}}$, positional encodings $X_{\text{PE}}$, adjacency matrix $A$, edge features $E_{\text{cont}}$}
\KwOut{Refined representation $X_{\text{out}}$}

Transform node features: $\hat{X}_{\text{node}} \gets L_{\text{node}}(X_{\text{UNI}})$\;
Transform edge features: $\hat{E}_{\text{edge}} \gets L_{\text{edge}}(E_{\text{cont}})$\;

\textbf{SlideGraph+ branch:} 
$X_{\text{SG}} \gets \text{SG\_Branch}(A, \hat{X}_{\text{node}}, \hat{E}_{\text{edge}})$\;

\textbf{Mamba branch:} 
$X_{\text{seq}} \gets \text{Concat}(\hat{X}_{\text{node}}, X_{\text{PE}})$\;
$X_{\text{Mamba}} \gets \text{Mamba\_Branch}(X_{\text{seq}})$\;

\textbf{Entropy-based fusion:}\;
$\hat{y}_{\text{SG}} \gets \text{Softmax}(X_{\text{SG}})$\;
$\hat{y}_{\text{Mamba}} \gets \text{Softmax}(X_{\text{Mamba}})$\;
$\alpha \gets \text{EntropyConfidence}(\hat{y}_{\text{SG}}, \hat{y}_{\text{Mamba}})$\;
$X_{\text{fused}} \gets (1-\alpha)\,X_{\text{SG}} + \alpha\,X_{\text{Mamba}}$\;

\textbf{Refinement:}\;
$X_{\text{mlp}} \gets \text{MLP}(X_{\text{fused}})$\;
$X_{\text{out}} \gets \text{BatchNorm}(X_{\text{mlp}} + X_{\text{fused}})$\;

\Return $X_{\text{out}}$\;
\end{algorithm}

\begin{algorithm}
\caption{EntropyConfidence}
\label{alg:EntropyConfidence}
\KwIn{Softmax outputs $\hat{y}_{\text{SG}}, \hat{y}_{\text{Mamba}}$; number of classes $C$}
\KwOut{Fusion weight $\alpha$}

Compute entropy for SG branch:\;
$H_{\text{SG}} \gets -\frac{1}{\log C} \sum_{c=1}^{C} \hat{y}_{\text{SG},c}\,\log(\hat{y}_{\text{SG},c})$\;

Compute entropy for Mamba branch:\;
$H_{\text{Mamba}} \gets -\frac{1}{\log C} \sum_{c=1}^{C} \hat{y}_{\text{Mamba},c}\,\log(\hat{y}_{\text{Mamba},c})$\;

Convert to confidence weights:\;
$w_{\text{SG}} \gets 1 - H_{\text{SG}}$\;
$w_{\text{Mamba}} \gets 1 - H_{\text{Mamba}}$\;

Normalize:\;
$\alpha \gets \dfrac{w_{\text{Mamba}}}{w_{\text{SG}} + w_{\text{Mamba}}}$\;

\Return $\alpha$\;
\end{algorithm}

\section*{Data}
To evaluate the effectiveness of SlideMamba, we conducted experiments using WSIs collected from the OAK clinical trial (NCT02008227)\cite{rittmeyer2017atezolizumab}. 
This dataset comprises 1,114 primary lung cancer cases, each with corresponding molecular profiling to establish ground-truth fusion and mutation status. The slides were scanned at 40× magnification, ensuring high-resolution representation of morphological features relevant to downstream computational pathology tasks.

For experimental design, the dataset was partitioned following a 5-fold cross-validation strategy. In each fold, cases were stratified into training (n = 561; 148 positive, 413 negative), validation (n = 312; 84 positive, 228 negative), and test (n = 223; 60 positive, 163 negative) cohorts, preserving the balance of positive and negative cases across splits. This design ensured that performance metrics reflect both robustness and generalizability.

The experimental task was formulated as a supervised slide-level classification problem, where the model predicts whether a given WSI corresponds to a sample with gene fusions or mutations. Predictions were evaluated against ground-truth molecular labels. To benchmark SlideMamba, we compared its performance against a set of competitive baselines, including Multiple Instance Learning (MIL), Transformer-based MIL (Trans-MIL), Mamba, SlideGraph+, and recently proposed GAT-Mamba\cite{ding2024combining}.

Performance was assessed using standard metrics: average precision (area under the precision–recall curve), area under the ROC curve (ROC AUC), sensitivity, and specificity. These metrics were reported as mean ± standard deviation across the five folds, enabling rigorous comparison of SlideMamba’s adaptive fusion strategy with established approaches.

\section*{Results}

As summarized in Table \ref{tab:results}, SlideMamba consistently outperformed all benchmark methods across multiple evaluation metrics. The model achieved a test average precision of $0.751 \pm 0.05$ and a test ROC AUC of $0.738 \pm 0.055$, indicating improved capacity to capture the complementary short- and long-range dependencies inherent in whole-slide image analysis. These results confirm the effectiveness of the proposed GNN–Mamba hybrid architecture with entropy-based adaptive fusion.

In terms of class-level performance, SlideMamba attained a sensitivity of $0.6625 \pm 0.083$ and a specificity of $0.725 \pm 0.094$. This balance demonstrates the ability of the model to reliably identify mutation-positive cases while controlling the false positive rate. Achieving both high sensitivity and specificity is essential in clinical applications, where accurate detection must be accompanied by the minimization of unnecessary follow-up procedures.

Comparisons with baseline models further demonstrate the benefits of the proposed approach. Traditional MIL and Trans-MIL methods yielded substantially lower average precision ($0.491 \pm 0.042$ and $0.39 \pm 0.017$, respectively), reflecting their limited ability to incorporate multiscale contextual information. Graph-based SlideGraph+ achieved a comparable average precision ($0.748 \pm 0.091$) but did not exceed SlideMamba in overall discriminative performance, highlighting the limitations of local-only modeling. Conversely, the Mamba-only branch ($0.664 \pm 0.063$) demonstrated inferior sensitivity, suggesting that reliance on global dependencies alone is insufficient for robust prediction.

Hybrid strategies such as GAT-Mamba provided moderate improvements (average precision $0.703 \pm 0.075$) but remained below the performance of SlideMamba. The observed gap supports the importance of adaptive, data-driven integration of local and global features. Unlike rigid element-wise fusion approaches, SlideMamba employs an entropy-guided mechanism that adjusts branch contributions dynamically, thereby enhancing predictive robustness across heterogeneous tissue morphologies.



\begin{table}[ht!]
\caption{Performance Comparison of SlideMamba Against Benchmark Models for Gene Mutation Detection. Values are reported as Mean $\pm$ Standard Deviation across 5-fold cross-validation.}
\centering
\begin{tabular}{|l|c|c|c|c|}
\hline
\textbf{Model} & \textbf{Test Average Precision} & \textbf{Test ROC AUC} & \textbf{Test Sensitivity} & \textbf{Test Specificity} \\
\hline
MIL & $0.491 \pm 0.042$ & $0.714 \pm 0.036$ & $0.4416 \pm 0.090$ & $854 \pm 0.0161$ \\
Trans-MIL & $0.39 \pm 0.017$ & $0.64 \pm 0.031$ & $0.297 \pm 0.019$ & $0.853 \pm 0.031$ \\
SlideGraph+ & $0.748 \pm 0.091$ & $0.733 \pm 0.085$ & $0.638 \pm 0.092$ & $0.75 \pm 0.04$ \\
GATMamba & $0.703 \pm 0.075$ & $0.723 \pm 0.07$ & $0.712 \pm 0.055$ & $0.762 \pm 0.12$ \\
Mamba & $0.664 \pm 0.063$ & $0.66 \pm 0.06$ & $0.475 \pm 0.071$ & $0.875 \pm 0.09$ \\
\hline
\textbf{SlideMamba} & $\mathbf{0.751 \pm 0.05}$ & $\mathbf{0.738 \pm 0.055}$ & $\mathbf{0.6625 \pm 0.083}$ & $\mathbf{0.725 \pm 0.094}$ \\
\hline
\end{tabular}
\label{tab:results}
\end{table}

\section*{Discussion}

In this work, we introduced SlideMamba, a hybrid framework that combines local relational modeling through a graph neural network branch with global context modeling via a Mamba-based state space branch. These two branches are further integrated using an entropy-guided adaptive weighting mechanism. Our experiments show that this design consistently outperforms unimodal baselines and existing hybrid strategies across multiple evaluation metrics. These results demonstrate the critical importance of effectively fusing multi-scale morphological information for challenging computational pathology tasks, including gene fusion and mutation prediction from H\&E images.


The core innovation of SlideMamba lies in its entropy-based adaptive fusion mechanism. Prior hybrid models, such as GAT-Mamba, have relied on rigid element-wise summation to merge local and global representations—a "one-size-fits-all" approach that overlooks the heterogeneity of whole-slide images (WSIs). In practice, the relative predictive value of local versus global features can vary widely across cases. For example, a WSI dominated by compact, distinctive tumor cell clusters may be best captured through the fine-grained relational modeling of a GNN, whereas a slide exhibiting diffuse architectural disorganization may benefit more from Mamba’s ability to capture long-range dependencies. 
SlideMamba addresses this challenge by dynamically weighting each branch’s contribution according to its predictive confidence. Specifically, the model computes the entropy of each branch’s softmax output, using lower entropy (higher confidence) as a signal to assign greater weight. When the GNN branch produces confident predictions from localized patterns, it dominates the fusion; conversely, when local patterns are ambiguous but global structure is clear, the Mamba branch is emphasized. This adaptive allocation ensures that the most reliable representation guides the final decision for each slide.
Beyond improving predictive accuracy, this mechanism mitigates a key limitation of summation-based fusion: the risk that noisy or uninformative features from one branch dilute the signals of the other. By making fusion context-aware and data-driven, SlideMamba provides a flexible and robust strategy for handling the intrinsic variability of WSIs.

Despite its promising performance, our study has several limitations that open avenues for future research. First, while our entropy-based fusion enhances interpretability by quantifying the relative reliance on local versus global features, the weights are static across the WSI, lacking the flexibility of spatially varying contributions. A promising direction for future work is to make the fusion weights ($\alpha$) patch-dependent and visualize them spatially across the WSI. Such a representation would highlight which tissue regions drive a preference for GNN versus Mamba, offering pathologists more intuitive insights into how the model integrates local and global information; Second, the current framework was evaluated on a single, albeit large, dataset for a specific task in lung cancer. To establish the generalizability of SlideMamba, it will be crucial to validate its performance across different cancer types (e.g., breast, colon, prostate) and for a wider range of downstream tasks, such as disease subtyping and survival prediction; Finally, our model's architecture, with its parallel branches, is computationally more intensive than a single-modality approach. Although Mamba's linear complexity offers significant efficiency gains over Transformers, further optimization may be needed to facilitate deployment in resource-constrained clinical settings. Exploring techniques like model pruning or knowledge distillation could be a promising direction.

In conclusion, SlideMamba effectively addresses a core challenge in WSI analysis by adaptively fusing local and global morphological features. By introducing an entropy-based confidence mechanism, our work moves beyond rigid fusion strategies and paves the way for more nuanced, context-aware, and ultimately more accurate multi-scale representation learning in digital pathology.

\section*{Acknowledgements}

We sincerely thank the support by the Computational Science and Informatics (CSI), Roche Diagnostics Solutions through the incubator project funding and internship program. We gratefully acknowledge the team at Analytics and Medical Imaging (AMI) , Genentech, Inc. / F. Hoffmann-La Roche AG for their valuable support in clinical image curation and outcome data collection. This work was supported by clinical trials funded by Genentech, Inc. / F. Hoffmann-La Roche AG. We extend our sincere gratitude to all patients and their families who made this research possible.

\section*{Funding}
This work did not receive any specific grant from funding agencies in the public, commercial, or not-for-profit sectors.

\section*{Declaration of interests}
F.D., N.S., Y.N., R.V. and X.L. are full-time employees of Roche and hold company stock.
The views expressed in this article are those of the authors and do not necessarily reflect the views of their affiliated institutions or employers.

\section*{Data availability}
For up-to-date details on Roche’s Global Policy on the Sharing of Clinical Information and how to request access to related clinical study documents, see here: \url{https://go.roche.com/data_sharing}. 

\bibliographystyle{plain} 
\bibliography{mybibliography}

\end{document}